\documentclass[conference]{IEEEtran}
\IEEEoverridecommandlockouts
\usepackage{cite}
\usepackage{amsmath,amssymb,amsfonts}
\usepackage{algorithmic}
\usepackage{graphicx}
\usepackage{textcomp}
\usepackage{xcolor}
\usepackage{subfigure}
\def\BibTeX{{\rm B\kern-.05em{\sc i\kern-.025em b}\kern-.08em
    T\kern-.1667em\lower.7ex\hbox{E}\kern-.125emX}}
\begin{document}

\title{Vehicle Trajectory Prediction by Transfer Learning of Semi-Supervised Models}

\makeatletter
\newcommand{\linebreakand}{%
  \end{@IEEEauthorhalign}
  \hfill\mbox{}\par
  \mbox{}\hfill\begin{@IEEEauthorhalign}
}
\makeatother

\author{\IEEEauthorblockN{Nick Lamm}
\IEEEauthorblockA{\textit{Department of Computer Science} \\
\textit{Columbia University}\\
New York, USA \\
nl2680@columbia.edu}
\and
\IEEEauthorblockN{Shashank Jaiprakash}
\IEEEauthorblockA{\textit{Department of Computer Science} \\
\textit{Columbia University}\\
New York, USA \\
sj3003@columbia.edu}
\and
\IEEEauthorblockN{Malavika Srikanth}
\IEEEauthorblockA{\textit{Department of Computer Science} \\
\textit{Columbia University}\\
New York, USA \\
ms5908@columbia.edu}
\linebreakand
\IEEEauthorblockN{Iddo Drori}
\IEEEauthorblockA{\textit{Department of Computer Science} \\
\textit{Columbia University}\\
New York, USA \\
idrori@cs.columbia.edu}}

\maketitle

\begin{abstract}
In this work we show that semi-supervised models for vehicle trajectory prediction significantly improve performance over supervised models on state-of-the-art real-world benchmarks. Moving from supervised to semi-supervised models allows scaling-up by using unlabeled data, increasing the number of images in pre-training from Millions to a Billion. We perform ablation studies comparing transfer learning of semi-supervised and supervised models while keeping all other factors equal. Within semi-supervised models we compare contrastive learning with teacher-student methods as well as networks predicting a small number of trajectories with networks predicting probabilities over a large trajectory set. Our results using both low-level and mid-level representations of the driving environment demonstrate the applicability of semi-supervised methods for real-world vehicle trajectory prediction.
\end{abstract}

\section{Introduction}

Predicting the trajectory of a vehicle in a multi-agent environment is a challenging and critical task for developing safe autonomous vehicles. State-of-the-art models rely on a representation of the environment from either direct, low-level input from sensors on the vehicle, or from a mid-level representation of the scene, which is commonly a map annotated with agent positions. Both of these approaches rely on a model to encode either camera data in the low-level case or annotated maps in the mid-level case. We show an example of both types of representations in Figure \ref{fig:io}. Mid-level representations as depicted in the top-left are used to predict candidate trajectories as shown in the top-right. Low-level representations such as camera data shown in the bottom-left can be used in an end-to-end fashion to predict steering angles as illustrated in the bottom-right. To encode these input representations, rather than training a model from scratch, state-of-the-art models rely on transfer learning with a model pre-trained on a supervised task \cite{messaoud2020trajectory, phan2020covernet} such as ImageNet classification. We perform an ablation study comparing transfer learning of supervised and semi-supervised models, while keeping all other factors equal, and show that semi-supervised models perform better than supervised models for both low-level and mid-level representations.

We demonstrate this comparison on state-of-the-art methods for vehicle trajectory prediction. For a low-level representation, we use the winning architecture of the ICCV 2019: Learning-to-Drive Challenge, which uses vehicle camera footage to predict the future speed and steering wheel angle \cite{diodato2019winning}. For a mid-level representation, we use CoverNet \cite{phan2020covernet} and multiple trajectory prediction (MTP) \cite{cui2019multimodal}, two multi-modal approaches that take an annotated map image as input. In all of these cases, we keep the architecture and computational resources the same, and compare semi-supervised and supervised models to encode the representation. Semi-supervised models have demonstrated state-of-the-art performance on computer vision benchmarks since they are able to learn from unlabeled datasets orders of magnitude larger than available labeled data \cite{caron2019unsupervised, he2020momentum, xie2020noisystudent, yalniz2019billionscale}. Notably, although annotated maps are not representative of the images in the datasets used to pre-train these models, they share common features with the mid-level map representation.

\begin{figure}
\centering
\includegraphics[width=0.5\textwidth]{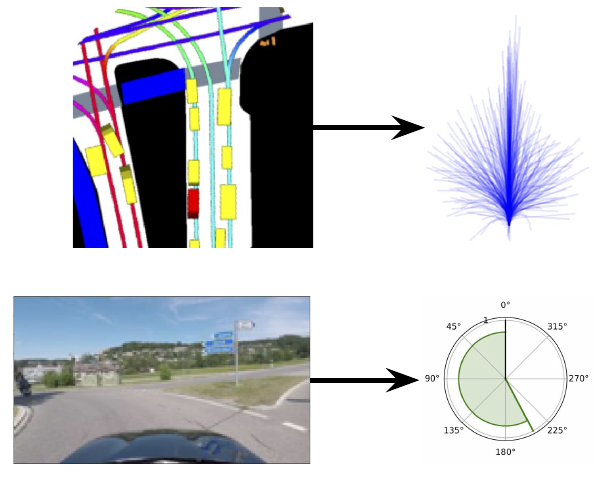}
\caption{An example of input and output representations for mid-level (top) and low-level representations (bottom). In the top row, the mid-level input representation is an annotated map of the scene (top left), with boxes representing agent positions and colors representing semantic categories. The output (top right) is a probability distribution over a set of candidate trajectories. In the bottom row, a low-level representation uses an image from the vehicle's front-facing camera as input (bottom left), and predicts the future steering wheel angle (bottom right) and speed of the vehicle.}
\label{fig:io}
\end{figure}

Our results demonstrate three key contributions for trajectory prediction (described in Section \ref{sec:results}):
\begin{enumerate}
    \item Semi-supervised models significantly improve upon supervised models using both low-level and mid-level representations as described in Section \ref{sec:results} and shown in Tables \ref{table:results} and \ref{table:drive360results}.
    \item Contrastive semi-supervised learning (SimCLR \cite{chen2020simclr}) outperforms teacher-student semi-supervised learning (ResNeXt-101 32x4d SSL and SWSL \cite{yalniz2019billionscale}).
    \item Using semi-supervised models for predicting probabilities of a large set of trajectories with CoverNet \cite{phan2020covernet} results in significant performance improvement over supervised models across both uni-modal and multi-modal metrics (up to 40.1\%); whereas using semi-supervised models for predicting a small set of trajectories with MTP \cite{cui2019multimodal} results in significant performance improvement only on uni-modal metrics (up to 17.3\%).
\end{enumerate}

\subsection{Related Work}

Low-level approaches to trajectory prediction use sensor data recorded by the vehicle, such as mounted cameras, as direct input to a model \cite{bojarski2016end,fernando2017going}. These approaches use a model to encode the raw pixels from the camera footage into a feature vector. We evaluate such a low-level representation \cite{diodato2019winning}, which uses front-facing camera images in combination with a vector of semantic map features to predict a vehicle's future steering wheel angle and speed.

Many approaches instead use a mid-level representation of the environment as input to the model \cite{chai2019multipath,cui2019multimodal,djuric-uncertainty,messaoud2020trajectory,phan2020covernet}. This commonly involves generating a map of the scene and annotating it with past and current positions of all other agents, using color to designate semantic categories of agents as well as static entities such as road boundaries and crosswalks. The map is then rasterized into an image, which serves as a compact mid-level representation of the entire scene. Similar to the low-level approach, the map image is fed through a model to generate a feature vector, which is used in a system of neural networks for trajectory prediction.

While systems often predict a single trajectory (mode), there is an advantage in predicting multiple modes and their associated probabilities, especially when there are multiple plausible trajectories that the vehicle might take. Several works \cite{chai2019multipath,cui2019multimodal,deo2018convolutional,messaoud2020trajectory,phan2020covernet,tang2019multiple,yang2018end} use a multi-modal approach, predicting a probability distribution over trajectories for agents in the environment. This approach has been extended using multi-head attention \cite{kim2020multi,messaoud2020trajectory}, allowing the model to focus on certain agents or other features of the scene context. In another approach \cite{yang2018end}, a multi-modal multi-task method jointly reasons about the future speed and steering of the vehicle, noting the joint relationship between the two. The Trajectron \cite{Trajectron} models multiple agents as dynamic graphs, and performs trajectory prediction for multi-modal, dynamic and variable multi-agent scenarios. SPAGNN \cite{casas2020spatiallyaware} addresses the behavior of other human drivers who make complex trade-offs while driving, modeling this relational behavior with graph neural networks.

Incorporating prior knowledge about the geometry and topology of roads into loss functions \cite{casas2020importance} has been shown to result in more precise trajectory distributions over future outcomes. Rules of the road \cite{hong2019rules} encodes high-level semantic information such as the entity state, other entities' states and road networks into a spatial grid allowing deep convolutional networks to learn entity-entity and entity-environment interactions. ChauffeurNet \cite{bansal2019chauffeurnet} introduces perturbations to trajectories and incorporates a loss for real-world driving mistakes, such as collisions and driving off-road. Our mid-level representation overlays multiple elements onto a single map for capturing the scene, losses, and driving goals.

Recent semi-supervised models extend and improve upon supervised ResNets by using orders of magnitude more unlabeled input samples \cite{chen2020simclr,he2020momentum,xie2020noisystudent,yalniz2019billionscale}, as well as output sample statistics \cite{lamm2020trajectograms}, with good results in real-world applications.

\section{Methods}

\begin{figure}
\centering
\includegraphics[width=0.3\textwidth]{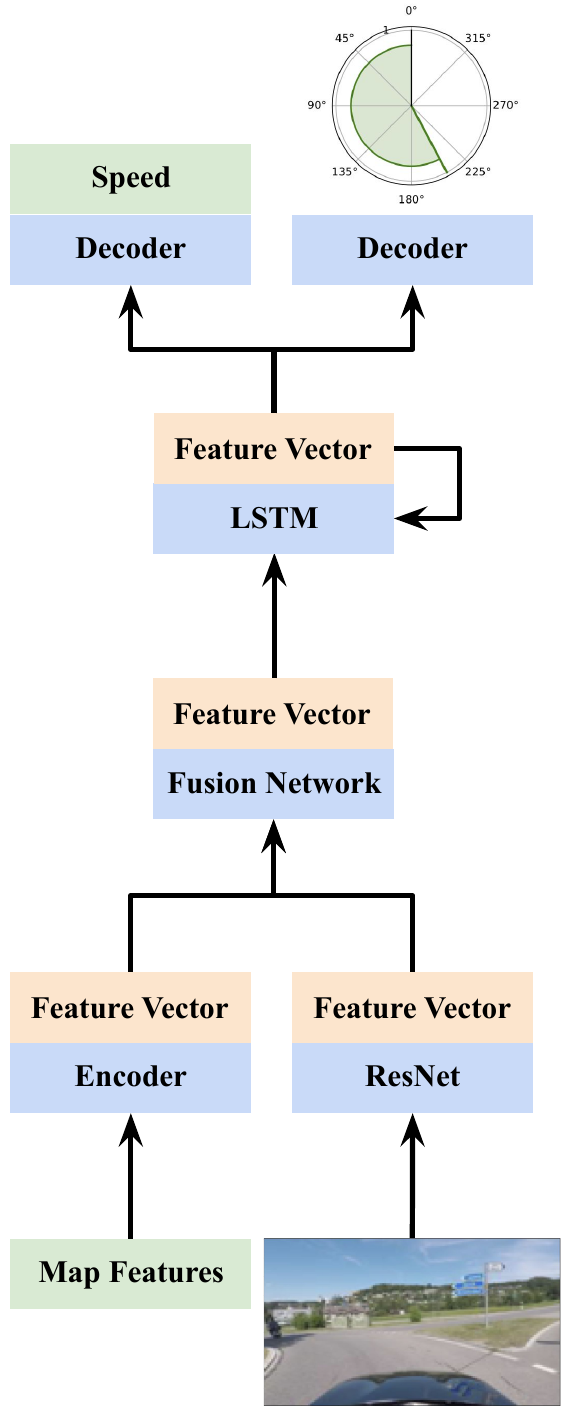}
\caption{System architectures for the low-level input representation. Inputs are shown at the bottom, neural networks in blue, intermediate feature vectors in orange, and the model output at the top. The network takes as input a vector of semantic map features and an image from the vehicle's front-facing camera. The inputs are encoded, and then fused together with a fully-connected network to capture non-linear interactions. An LSTM combines observations from multiple timesteps, and finally decoder networks predict the speed and the steering wheel angle of the driver.}
\label{fig:low-level-architecture}
\end{figure}

\begin{figure}
\centering
\includegraphics[width=0.3\textwidth]{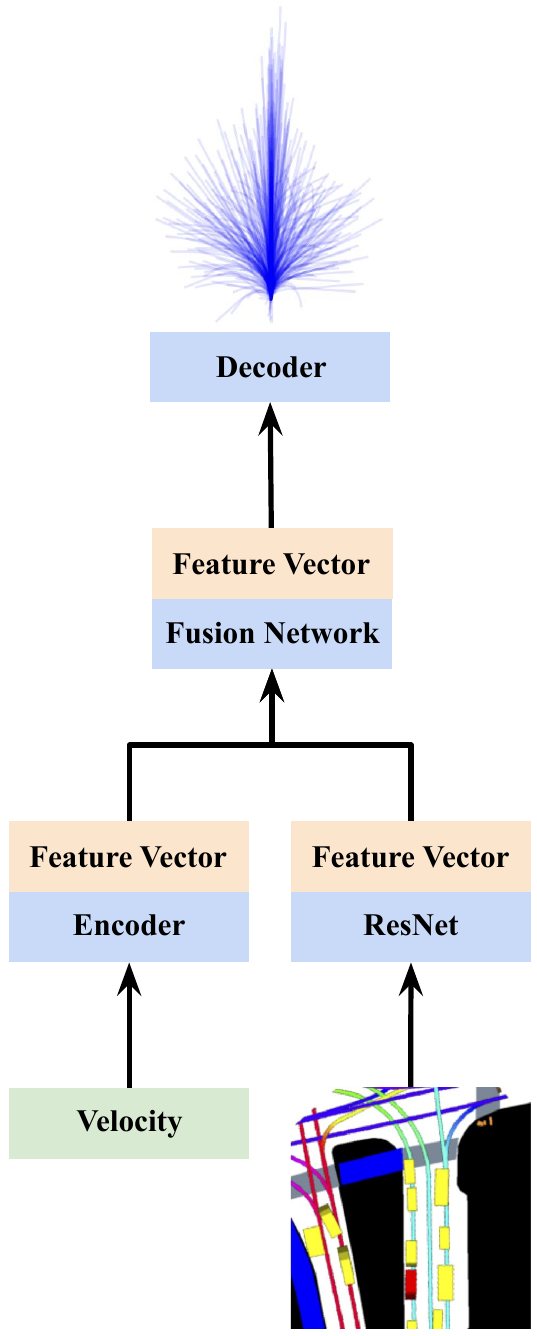}
\caption{System architectures for the mid-level input representation. Inputs are shown at the bottom, neural networks in blue, intermediate feature vectors in orange, and the model output at the top. The network takes as input the current state of the agent (a vector including velocity, acceleration, and yaw change rate), and an annotated map image of the environment. The map is fed into a ResNet backbone, and this representation is then concatenated with the agent state vector and passed to a fully-connected fusion network. For MTP, the final layer outputs a set of trajectories and an associated probability distribution. For CoverNet we use a fixed setting in which the output is a probability distribution over a set of candidate trajectories.}
\label{fig:mid-level-architecture}
\end{figure}

We perform ablation studies comparing transfer learning of semi-supervised and supervised models on trajectory prediction tasks. We examine both low-level representations, which use the vehicle's front-facing camera images as input, and mid-level representations, which use an annotated map image as input. In both cases, we use different semi-supervised models to encode the input, while keeping all other factors equal, including the system architecture and computational resources.

\subsection{Input Representation}

\paragraph{Mid-level representation}
Following state-of-the-art trajectory prediction models \cite{chai2019multipath,cui2019multimodal,djuric-uncertainty,messaoud2020trajectory,phan2020covernet}, we generate an annotated map image to represent the driving environment. This includes annotations for drivable areas, crosswalks and walkways using color coding to represent semantic categories. All scenes are oriented such that the agent under consideration is centered and directed towards the top of the image. The positions of all agents in the scene are drawn onto the image, using faded bounding boxes to represent past positions in a historical window. By encoding all this information into a single map, a large amount of information is condensed into a single image. The top row of Figure \ref{fig:io} shows an annotated map of a scene in the nuScenes dataset \cite{nuscenes}. In addition to the map, a vector of the target agent's state at the moment of prediction is also included as input. This includes the agent's speed (between 0 and 30 m/s), acceleration (between -25 and 25 m/s$^2$) and yaw rate (between $-2\pi$ and $2\pi$ radians/s).

\paragraph{Low-level representation}
We use front-facing camera images from the Drive360 dataset \cite{drive360} as a low-level representation of a driving environment. In addition to the image, we include a vector of semantic map data, which includes datapoints such as the distance to the nearest intersection, the speed limit, and the approximate road curvature. The bottom row of Figure \ref{fig:io} show an example image from a front-facing camera in the Drive360 dataset.

\subsection{Semi-Supervised Models}
State-of-the-art models \cite{messaoud2020trajectory,phan2020covernet} use transfer learning of supervised models, whereas we evaluate the use of semi-supervised models. We perform transfer learning by fine-tuning each semi-supervised model on our training set, leveraging models already trained on up to a Billion images, orders of magnitude larger than the nuScenes dataset \cite{nuscenes} which consists of 1.4 Million images. We provide a summary of the semi-supervised models we use in Table \ref{table:semi-supervised-comparison}. Next, we describe each semi-supervised model in detail.

\paragraph{Teacher-student self-training} We use ResNeXt-101 32x4d SSL and SWSL \cite{yalniz2019billionscale}. ResNeXt-101 32x4d SSL is trained on a semi-supervised task using a teacher-student method on an unlabeled dataset of 90M images, and fine-tuned on 1.2M images from the ImageNet1k dataset. ResNeXt-101 32x4d SWSL is trained using a teacher-student method on 940M images, leveraging associated hashtags in a semi-weakly supervised approach, and fine-tuned on the ImageNet1k dataset. Both of these models use the ResNeXt-101 32x4d architecture from \cite{resnext}.

\paragraph{Contrastive learning} We use SimCLR \cite{chen2020simclr}, trained using a contrastive learning method on ImageNet1k. During training, augmented versions of images are passed through a ResNet architecture \cite{resnet}. The contrastive loss objective serves to minimize the distance between different augmentations of the same image, and maximize the distance between representations of other images. We use a ResNet-50 architecture trained with the SimCLR method.

\setlength{\tabcolsep}{4pt}
\begin{table*}
\begin{center}
\caption{
Comparison of semi-supervised models used in our experiments. The labeled dataset in all the models consists of 1.2M images. Since SimCLR is trained on augmentations, there is no measure of unlabeled dataset size.}
\label{table:semi-supervised-comparison}
\begin{tabular}{l|l|l|l|l}
Model & Size &	Type & Label Ratio & Parameters\\
    \hline
ResNeXt-101 32x4d SWSL
& 940M
 & Teacher-student

& 1:780
& 42M \\

    ResNeXt-101 32x4d SSL
    & 90M
    & Teacher-student
    & 1:75
    & 42M
    \\

    SimCLR ResNet-50
    & N/A
    & Contrastive learning
    & N/A
    & 25.6M
    \\    \hline
  \end{tabular}
\end{center}
\end{table*}
\setlength{\tabcolsep}{1.4pt}

\subsection{Experiments}

We experiment with using transfer learning from semi-supervised models in place of supervised models for both low-level and mid-level representations of the input.

\paragraph{Mid-level representation}
For mid-level representations, we train our models to predict a 6-second trajectory for an agent, using 2 seconds of historical observations of the scene represented as an annotated map image. We use two architectures that have been shown to be successful on this task: Multiple-Trajectory Prediction (MTP) \cite{cui2019multimodal} and CoverNet \cite{phan2020covernet}, which we describe in more detail below. For each architecture, we substitute different semi-supervised models for the backbone used to encode the map image.

\paragraph{Low-level representation}
For low-level representations, we test our approach by training models on the Drive360 dataset \cite{drive360} used in the ICCV 2019: Learning-to-Drive Challenge. The task is to predict the speed and steering wheel angle of a human driver one second in the future after the observation. We experiment with different semi-supervised and supervised models to encode the front-facing camera footage, analogous to our experiments with the input map image of the mid-level representation. We use the architecture of the winning team of the competition \cite{diodato2019winning} which uses a supervised image encoder and has been shown to be an effective end-to-end model.

\subsection{Datasets}
\paragraph{nuScenes}
For our experiments with mid-level representations, we use nuScenes \cite{nuscenes}, a public large-scale dataset which consists of 1000 driving scenes in Boston and Singapore. Each scene is 20 seconds in length and is sampled at a frequency of 2Hz. We use the official data partitions from the nuScenes prediction challenge: 32,186 instances in the training set, 8,560 in the validation set, and 9,041 in the test set. Each instance is comprised of a scene at a particular point in time, with a particular agent of interest whose trajectory the model predicts. The dataset includes a high definition map of the scene, bounding boxes and past positions for all agents.

\paragraph{Drive360}
For our experiments with low-level representations, we use the Drive360 dataset \cite{drive360}. The dataset includes 55 hours of driving recorded in Switzerland, divided into 27 routes and 682 chapters. We partition the data into disjoint datasets for training (43\%), validation (43\%), and test (14\%). The dataset contains observations at a frequency of 10Hz, including GoPro images positioned around the car, of which we only use the front-facing camera, and map features in the form of a vector with 20 semantic datapoints such as the distance to the nearest intersection, the current speed limit, and the road curvature.

\subsection{Architectures}
We perform our experiments with low-level representations on the winning architectures of the ICCV 2019: Learning-to-Drive challenge \cite{diodato2019winning}, as shown in Figure \ref{fig:low-level-architecture}, trained on the Drive360 dataset. The architecture for mid-level representations is shown in Figure \ref{fig:mid-level-architecture}. We use two networks that are successful on the nuScenes dataset: (i) Multiple-Trajectory Prediction (MTP) \cite{cui2019multimodal} which predicts a small number of trajectories; and (ii) CoverNet \cite{phan2020covernet} which assigns probabilities to a large set of trajectories. In all cases, we hold constant the configuration of the architecture during all experiments, and vary the ResNet component used to encode the images with different semi-supervised and supervised models.

\paragraph{Mid-level representation}
MTP \cite{cui2019multimodal,phan2020covernet} uses the annotated map image and the target agent's current state to predict a fixed number of trajectories, as well as their associated probabilities. The map image is passed through the ``backbone" vision component, which is the model that we vary in our experiments. This representation and a vector of the agent's state are passed through a fully-connected neural network used for fusing the different inputs. The output is a set of trajectories $\mathcal{K}$, and a vector of logits corresponding to their probabilities. The loss is calculated as a sum of the classification loss $L_C$, which is a cross-entropy with the positive sample determined by the element in the trajectory set closest to the ground truth, referred to as the ``best matching" mode, and a regression loss $L_R$ for the best matching mode and the ground truth. In our experiments, we fix the number of output trajectories to 3. This matches one of the configurations evaluated in \cite{phan2020covernet}.

CoverNet \cite{phan2020covernet} performs trajectory prediction by computing the probability distribution over a set of candidate trajectories. Similar to MTP, the model uses the annotated map image and a vector representing the target agent's state as input. However, rather than predicting an entire set of trajectories $\mathcal{K}$ and their associated probabilities, the model only outputs probabilities for a fixed trajectory set $\mathcal{K}$. Although the original paper evaluates these scenarios using a dynamic and hybrid version of this trajectory set, we use the fixed version provided in the nuScenes dev-kit implementation for all our experiments.
The loss function is only the classification loss $L_C$ of the closest trajectory to the ground truth. In our experiments, we use the set of 415 trajectories. We show a visualization of this trajectory set in the top-right of Figure \ref{fig:io}.

\paragraph{Low-level representation}
The architecture for the low-level representation is depicted in Figure \ref{fig:low-level-architecture}. The model consists of neural networks (in blue) and intermediate feature vectors (in orange). We replace the ResNet with different semi-supervised and supervised models in our experiments. Images are fed into the ResNet model and the vector of semantic map data is passed through an encoder. These are then fused together using a fusion layer to capture the non-linear interactions between the data sources. An LSTM then combines observations from the current timestep and a recent timestep (400ms in the past). This output is then fused together with data from the initial timestep and passed through regressors to obtain the vehicle speed and steering angle prediction which are shown in green. The overall loss is the sum of the regression losses for the two targets.

\section{Results}
\label{sec:results}

\paragraph{Mid-level representation}
We perform experiments showing the performance of transfer learning from semi-supervised models for encoding annotated maps in a 6-second trajectory prediction task. We use two architectures: CoverNet and MTP. For each semi-supervised model, we compare against a supervised model trained on ImageNet with the same architecture and number of layers. We additionally include SimCLR with the wider ResNet-50(4x) \cite{chen2020simclr} architecture, one of the latest and best performing semi-supervised models on ImageNet benchmarks to date, to evaluate how improvements in semi-supervised pre-training contribute to our task.

We compare CoverNet and MTP models by a standard set of metrics for multi-modal trajectory prediction: minADE$_{1}$, minADE$_{5}$, minADE$_{10}$, FDE and HitRate$_{5,2m}$. The minimum Average Displacement Error (minADE$_{k}$) is the minimum displacement of the $k$ most likely trajectories from the ground truth, averaged along corresponding points of the ground truth and predicted trajectories. The HitRate$_{k,d}$ \cite{phan2020covernet} is the average number of trajectory sets in which this minimum, maximised along corresponding points of the ground truth and predicted trajectories, is below a threshold $d$. The final displacement error (FDE) is the error between the final predicted point and ground truth trajectory position, for the most likely trajectory. minADE$_{5}$, minADE$_{10}$ and HitRate$_{5,2m}$ take into consideration multiple modes while the other metrics are uni-modal.

As shown in Table \ref{table:results}, using semi-supervised models instead of supervised models shows significant improvement on most metrics when all other factors are held equal. Semi-supervised models result in minADE$_{1}$ improvements ranging from 5.8\% to 33.9\%, minADE$_{5}$ improvements up to 17.8\% and minADE$_{10}$ improvements as high as 15.5\% across CoverNet and MTP. The improvement in FDE from semi-supervised models are as high as 28.8\%. The improvement in HitRate$_{5,2m}$ is as high as 33\% when SimCLR Resnet-50 replaces supervised ResNet-50 in the CoverNet architecture.

It is notable that SimCLR ResNet-50(4x) outperforms all other semi-supervised models under consideration for the CoverNet architecture. SimCLR ResNet-50(4x) is relatively new and known to be one of the best performing semi-supervised models on the ImageNet dataset. This shows that improvements in semi-supervised pre-training can be leveraged to improve results in this domain through transfer learning.

\setlength{\tabcolsep}{4pt}
\begin{table*}
\begin{center}
\caption{Results of CoverNet and MTP on the nuScenes dataset, comparing different semi-supervised and supervised models to encode the annotated map. For each semi-supervised model, we make a direct comparison to a supervised model with the same architecture. Semi-supervised models significantly outperform their supervised counterparts on most metrics. Additionally we experiment with SimCLR ResNet-50(4x), one of the latest and top performing semi-supervised models to see how improvements in semi-supervised pre-training contribute to our task. For minADE(mADE) and FDE, lower is better, and for HitRate(HR) higher is better.\\}
\label{table:results}
\begin{tabular}{l|c|c|c|c|c|c}
\textbf{Model} & \textbf{Type} & \textbf{mADE\textsubscript{1}} & \textbf{mADE\textsubscript{5}} &	\textbf{mADE\textsubscript{10}} & \textbf{FDE} & \textbf{HR\textsubscript{5,2m}}\\
\hline
\textbf{Baselines} & & & & & &\\
\hline
Constant velocity
&
& 5.48
& 5.48
& 5.48
& 13.44
& 0.05
\\
Physics oracle
&
& 3.91
& 3.91
& 3.91
& 9.53
& 0.10
\\
\hline \hline
\textbf{CoverNet} & & & & & &\\ \hline
ResNet-50
& Supervised
& 9.23
& 3.03
& 2.20
& 18.48
& 0.12
\\
SimCLR ResNet-50
& Semi-Supervised
&6.10
&\textbf{2.49}
&1.86
&13.16
&0.16
\\
SimCLR ResNet-50(4x)
& Semi-Supervised
& \textbf{5.53}
& 2.52
& \textbf{1.86}
& \textbf{11.95}
& \textbf{0.16}
\\
ResNeXt-101 32x4d
& Semi-Supervised
& 9.28
& 2.95
& 2.10
& 18.75
& 0.14
\\
ResNeXt-101 32x4d SSL
& Semi-Supervised
& 7.03
& 2.67
& 1.99
& 14.67
& 0.14
\\
ResNeXt-101 32x4d SWSL
& Semi-Weakly Super.
& 7.43
& 2.65
& 1.99
& 16.64
& 0.14
\\
\hline \hline
\textbf{MTP} & & & & & &\\ \hline
ResNet-50
& Supervised
& 5.13
& 2.97
& 2.97
& 11.71
& 0.14
\\
SimCLR ResNet-50
& Semi-Supervised
&4.83
&3.04
&3.04
&11.11
&0.14
\\
SimCLR ResNet-50(4x)
& Semi-Supervised
& \textbf{4.69}
& 3.13
& 3.13
& \textbf{10.65}
& 0.11
\\
ResNeXt-101 32x4d
& Supervised
& 6.26
& 2.98
& 2.98
& 13.93
& 0.13
\\
ResNeXt-101 32x4d SSL
& Semi-Supervised
& 6.02
& 3.06
& 3.06
& 13.50
& 0.13
\\
ResNeXt-101 32x4d SWSL
& Semi-Weakly Super.
& 5.18
& \textbf{2.96}
& \textbf{2.96}
& 11.63
& \textbf{0.15}
\\
\hline
\end{tabular}
\end{center}
\end{table*}
\setlength{\tabcolsep}{1.4pt}

In Figure \ref{fig:hitrate}, we show the 2-meter HitRate metric as we increase $k$, the number of most probable trajectories included in the metric, for our experiments with CoverNet. It is clear that even over a wide range of $k$, the semi-supervised models outperform the supervised models, with SimCLR performing the strongest. We note that the supervised ResNet-50 model is a popular backbone model used in several implementations of CoverNet \cite{messaoud2020trajectory,phan2020covernet}, and our SimCLR model shows a clear improvement over this on all metrics without increasing the number of layers or inference time. Of all the semi-supervised methods, SimCLR, trained with constrastive learning, outperforms ResNeXt-101 SSL and SWSL, both trained with noisy-student methods.

We notice that while semi-supervised models perform better than supervised models across all metrics on CoverNet, this is not the case for MTP. For MTP, semi-supervised models improve performance significantly on the uni-modal metrics, however they perform only incrementally better or worse than supervised models on the multimodal metrics. This can be attributed to the fact that MTP predicts a small set of modes (3), as opposed to CoverNet which assigns probabilities to a much larger set of modes (415).

Examples of mid-level representations from the nuScenes dataset with their corresponding trajectory predictions using CoverNet are illustrated in Figure \ref{fig:sampleCoverNet}, and predictions using MTP are shown in Figure \ref{fig:sampleMTP}.

\begin{figure}
\centering
\includegraphics[width=0.4\textwidth]{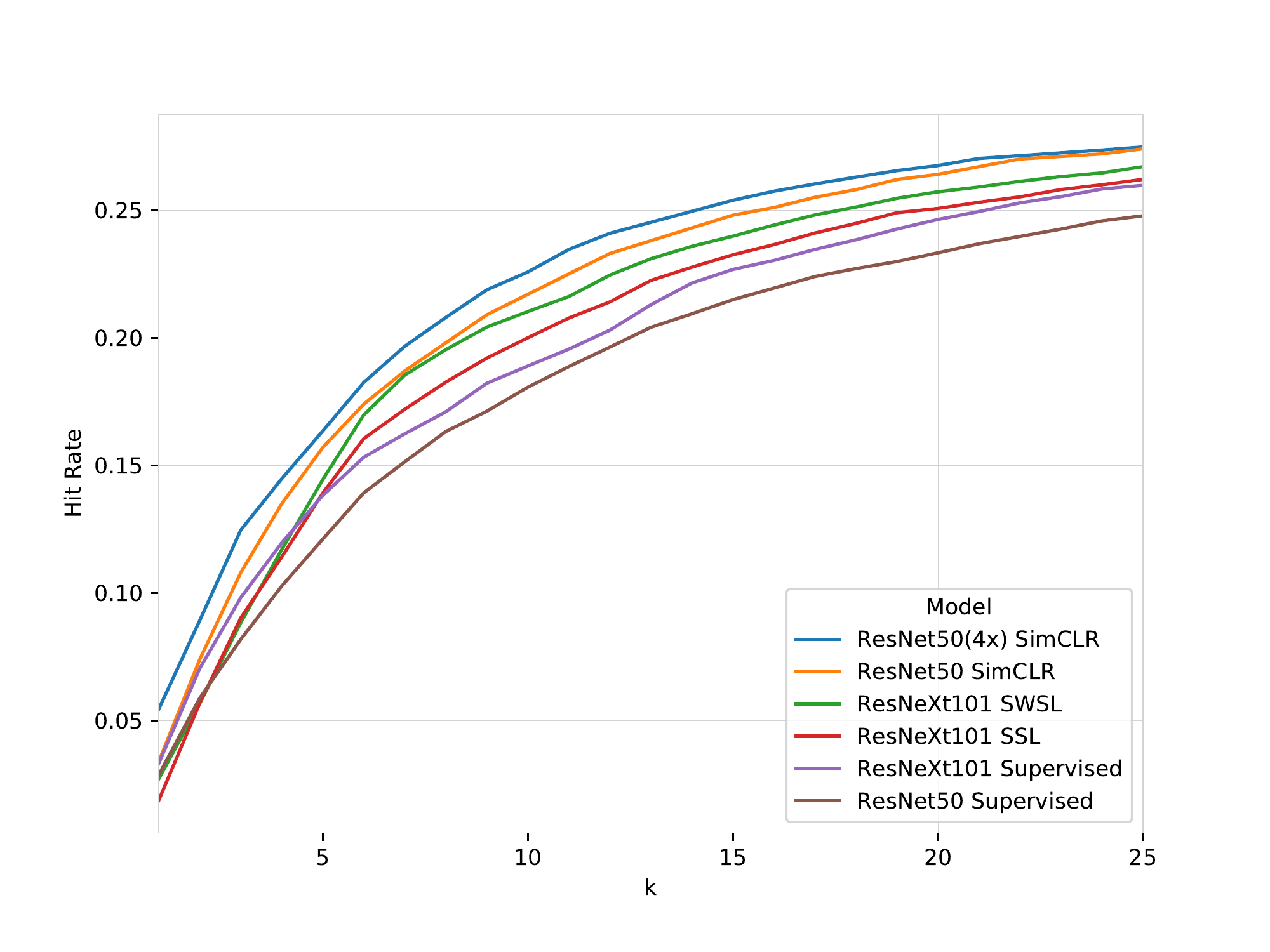}
\caption{Hit Rate for CoverNet (415 modes, fixed) for each backbone model over the top k predicted trajectories as k is increased. Beginning around k=5, there is a clear separation between the different backbone models, with the semi-supervised models outperforming the supervised models. As k increases, the relative ordering of the models remains for the most part constant. This indicates that increasing the number of candidate trajectories considered in the Hit Rate metric has a consistent effect across all the models.}
\label{fig:hitrate}
\end{figure}

\begin{figure}
\centering

\subfigure{
\includegraphics[width=0.45\textwidth]{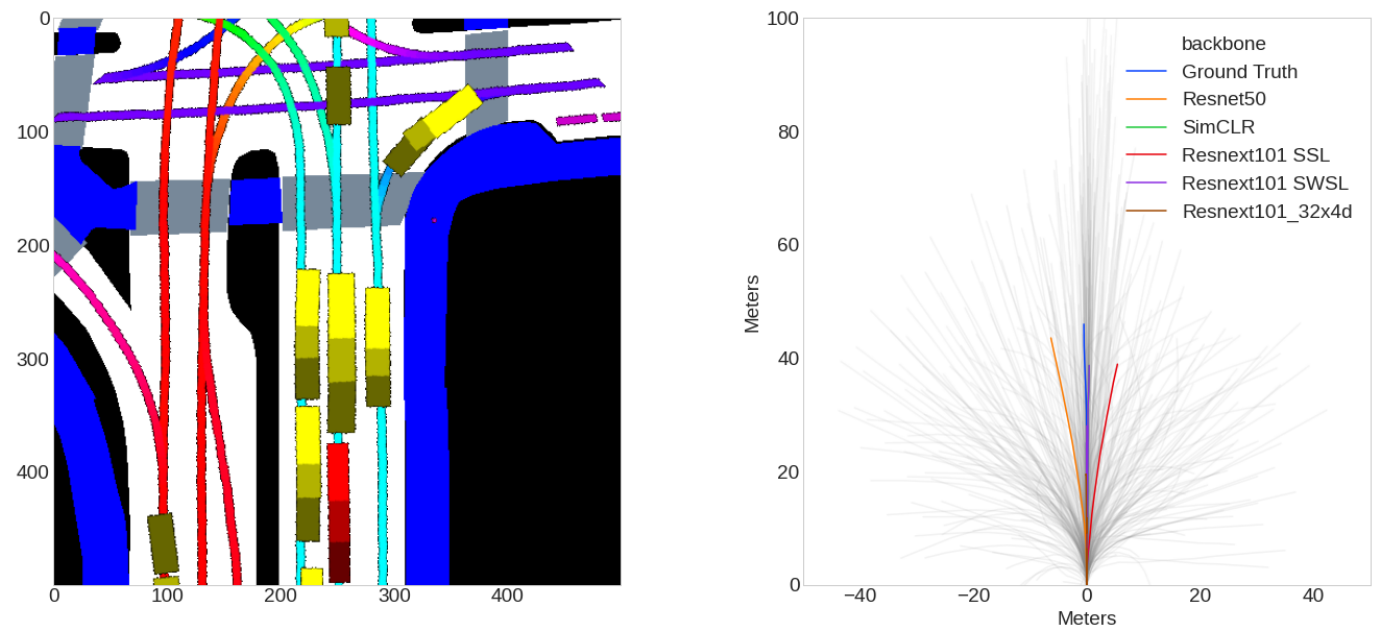}}

\subfigure{
\includegraphics[width=0.45\textwidth]{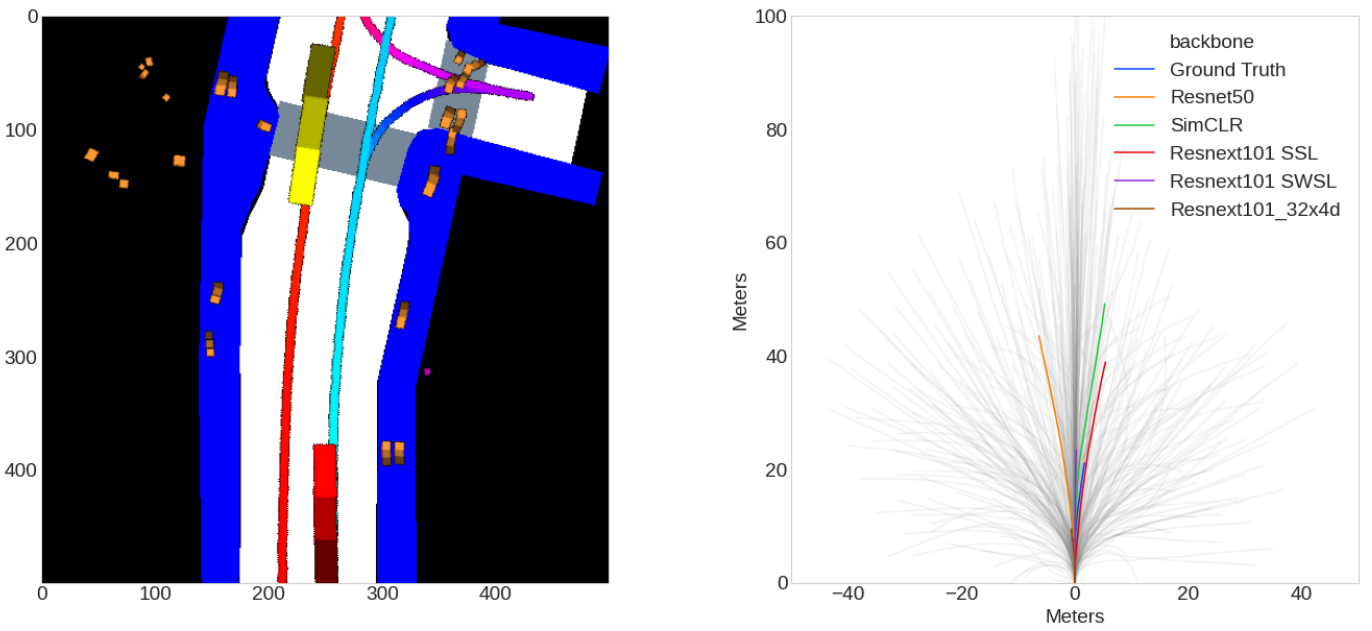}}

\subfigure{
\includegraphics[width=0.45\textwidth]{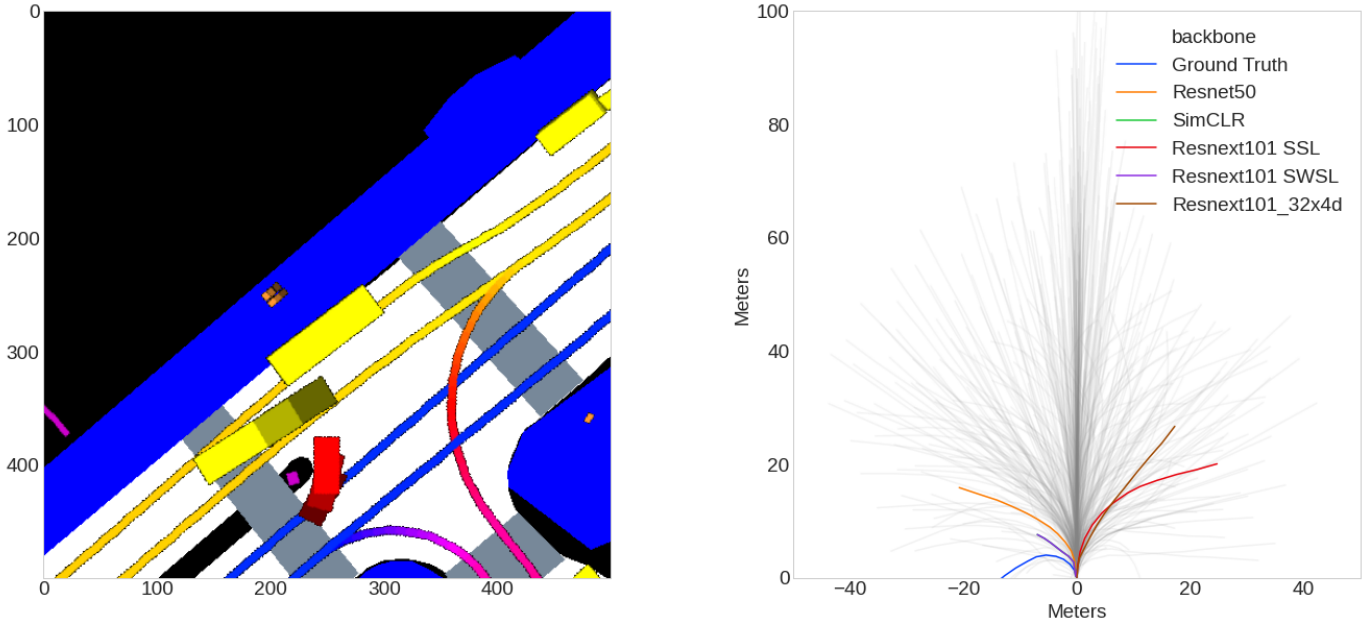}}

\caption{Trajectory prediction examples on the nuScenes dataset using CoverNet. The mid-level representation of the annotated map is on the left. The colored lines on the right represent the ground truth trajectory (blue) and those predicted by CoverNet with various backbone models, including the supervised ResNet-50 (orange) and the semi-supervised models we evaluate. The gray lines in the background are the set of 415 trajectories in the fixed trajectory set.}

\label{fig:sampleCoverNet}
\end{figure}

\begin{figure}
\centering

\subfigure{
\includegraphics[width=0.45\textwidth]{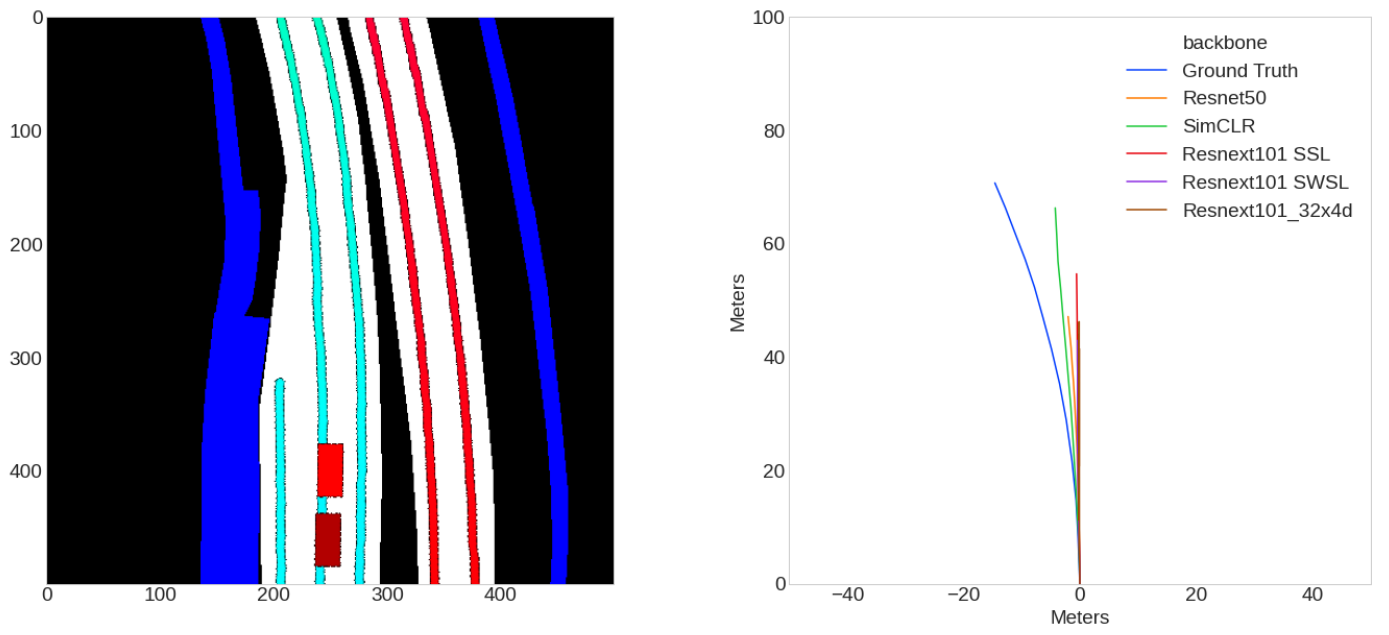}}

\subfigure{
\includegraphics[width=0.45\textwidth]{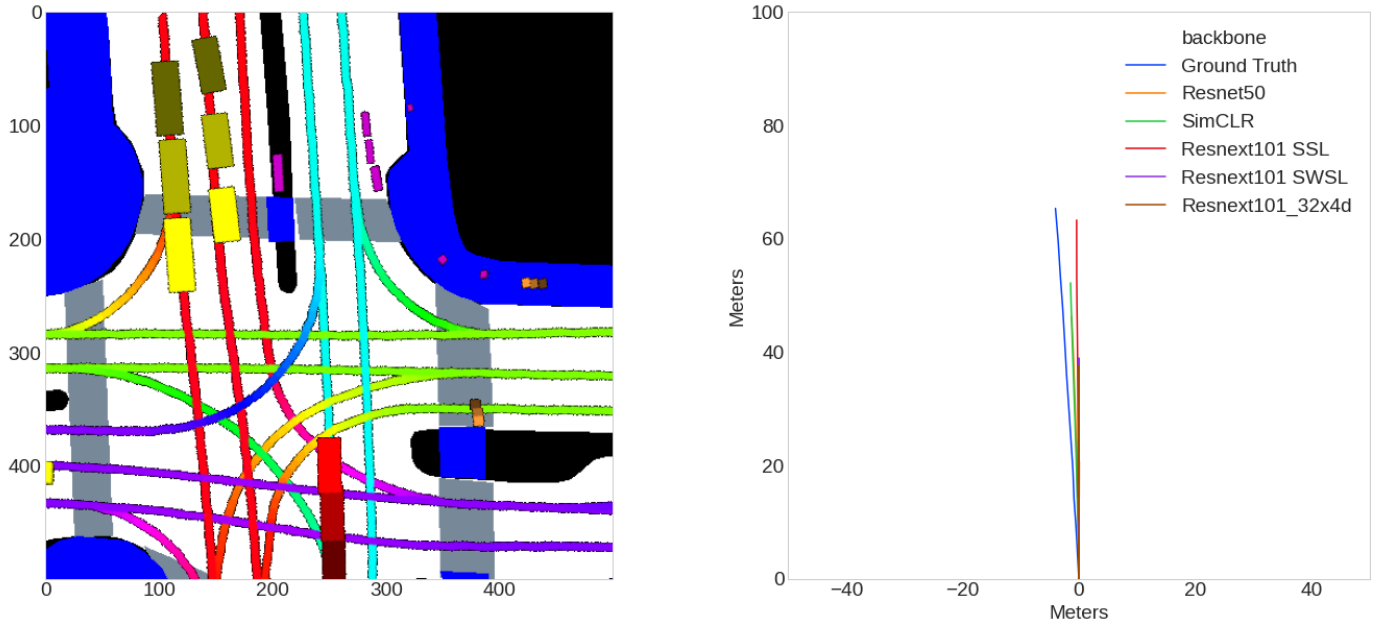}}

\subfigure{
\includegraphics[width=0.45\textwidth]{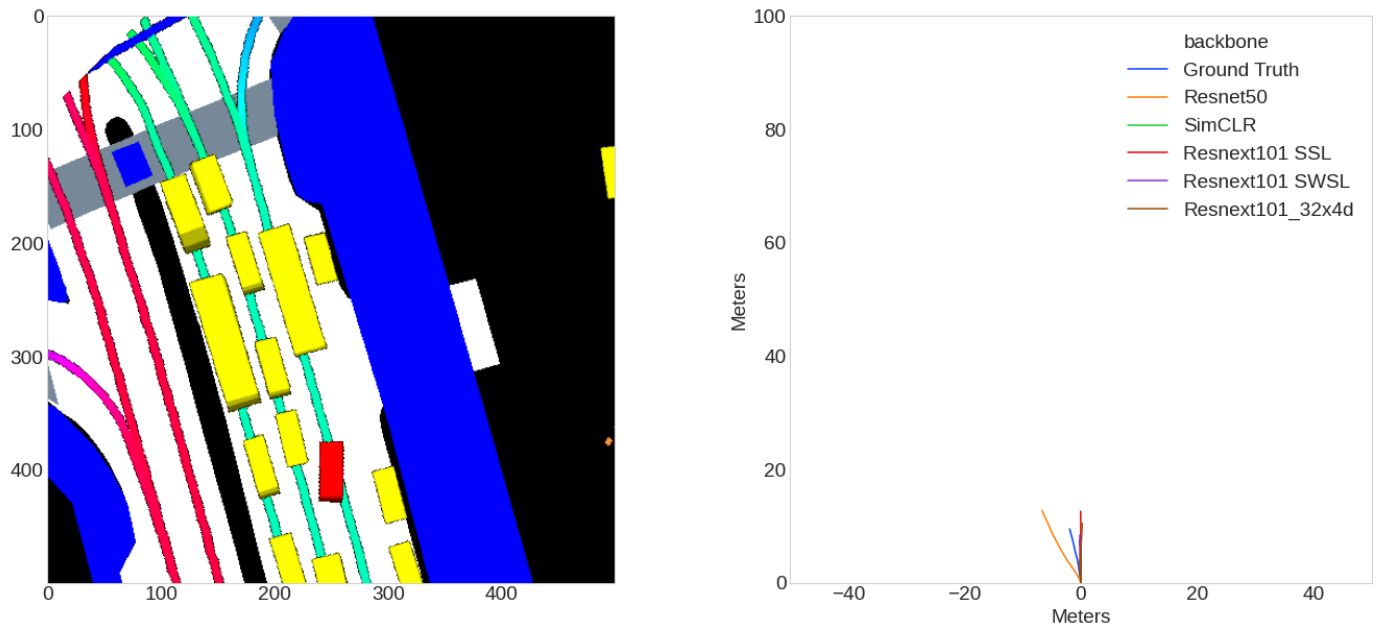}}

\caption{Trajectory prediction examples on the nuScenes dataset using MTP. The mid-level representation of the annotated map is on the left. The colored lines on the right represent the ground truth trajectory (blue) and those predicted by MTP with various backbone models, including the supervised ResNet-50 (orange) and the semi-supervised models we evaluate. In the third example, we show a dense traffic scenario.}

\label{fig:sampleMTP}
\end{figure}

\paragraph{Low-level representation}
The results of our experiments on low-level representations using the Drive360 dataset are shown in Table \ref{table:drive360results}. For the ICCV 2019: Learning-to-Drive winning architecture (L2D), which predicts the speed and steering for a timestep one second in the future, we report mean squared error (MSE) for both targets. SimCLR performs the best on the overall dataset, having the lowest MSE for both speed and steering wheel angle, outperforming the supervised models. This reiterates the findings from our experiments on mid-level representations where SimCLR, trained with constrastive learning, outperforms the other models in most cases. We however do not observe improvements when using the semi-supervised ResNext-101 32x4d SSL and ResNext-101 32x4d SWSL models, trained with noisy-student methods, as compared to the supervised ResNet-101 on this task.

Examples of low-level inputs from the Drive360 dataset and their corresponding steering wheel angle predictions are shown in Figure \ref{fig:sampleDrive360}.

\setlength{\tabcolsep}{2pt}
\begin{table}
\begin{center}
\caption{
Comparison of speed and steering wheel angle prediction on the Drive360 test dataset for the semi-supervised and supervised models we evaluate. For both speed and steering wheel angle prediction, the semi-supervised SimCLR model improves upon the supervised models. Steering angle MSE is measured in $\textrm{degrees}^2$ and the speed MSE in $(\textrm{km}/\textrm{h})^2$.\\}
\label{table:drive360results}
\begin{tabular}{l|l||l|l}
\textbf{Model} & \textbf{Type} & \textbf{Angle MSE} & \textbf{Speed MSE}\\
\hline
\textbf{L2D winner on Drive360} & &\\
\hline
ResNet-50
& Supervised
& 1013.46
& 10.40
\\
SimCLR ResNet-50
& Semi-Supervised
& \textbf{1003.56}
& \textbf{9.53}
\\
ResNet-101
& Supervised
& 1010.64
& 10.43
\\
ResNeXt-101 32x4d SSL
& Semi-Supervised
& 1050.58
& 10.80
\\
ResNeXt-101 32x4d SWSL
& Semi-Weakly Super.
& 1103.13
& 9.69
\\
\hline
  \end{tabular}
  \end{center}
\end{table}
\setlength{\tabcolsep}{1.4pt}

\paragraph{Implementation Details}
Training is performed on a Google Cloud Platform instance with an NVIDIA Tesla T4 or P100 GPU. For the mid-level representations, we downsample the nuScenes training data by a ratio of 5:1 during training, which reduces training time to 10-20 hours per model. For the low-level representations, we downsample the Drive360 dataset by a ratio of 10:1 during training to reduce the number of training instances, and we additionally downsample the input images from from 1920x1080 to 160x90 pixels. This reduces training to about 5-10 hours per model. For all models, we report results on the complete test split without downsampling. During training, we freeze $\frac{3}{4}$ of the lowest blocks of the semi-supervised and supervised models, fine-tuning the remaining blocks.

\section{Conclusion}

We demonstrate the benefits of using transfer learning of semi-supervised models on real-world driving benchmarks. By performing an ablation study comparing transfer learning of semi-supervised models with supervised models while keeping all other factors equal, we show that using semi-supervised models improves performance for both low-level and mid-level representations. Within semi-supervised models, we compare: (i) contrastive learning with teacher-student methods; and (ii) networks predicting a small number of trajectories with networks predicting the probabilities over a large set of trajectories. Using semi-supervised models in place of supervised models requires no additional computational resources when performing transfer learning or inference, hence our results present a simple recipe for significantly improving trajectory prediction.

\clearpage

\begin{figure}
\centering

\subfigure{
\includegraphics[width=0.5\textwidth]{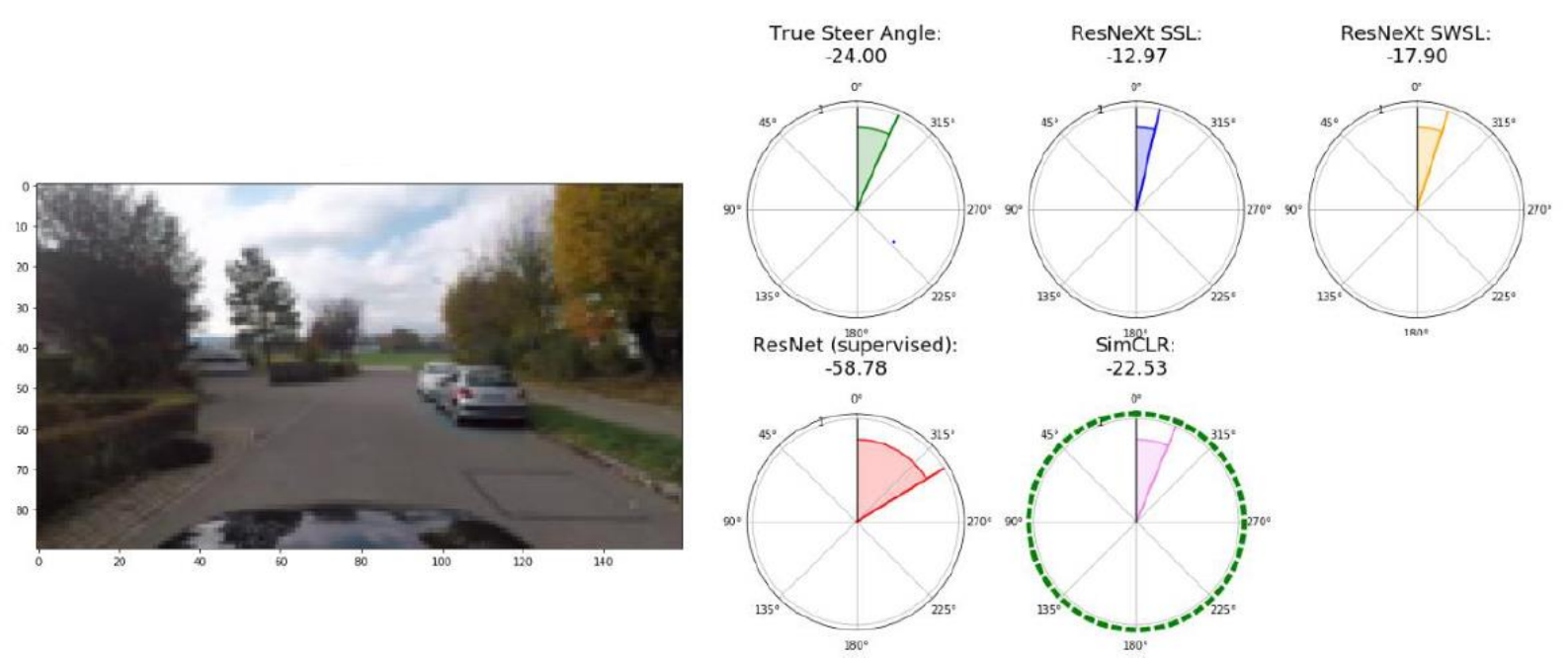}}

\subfigure{
\includegraphics[width=0.5\textwidth]{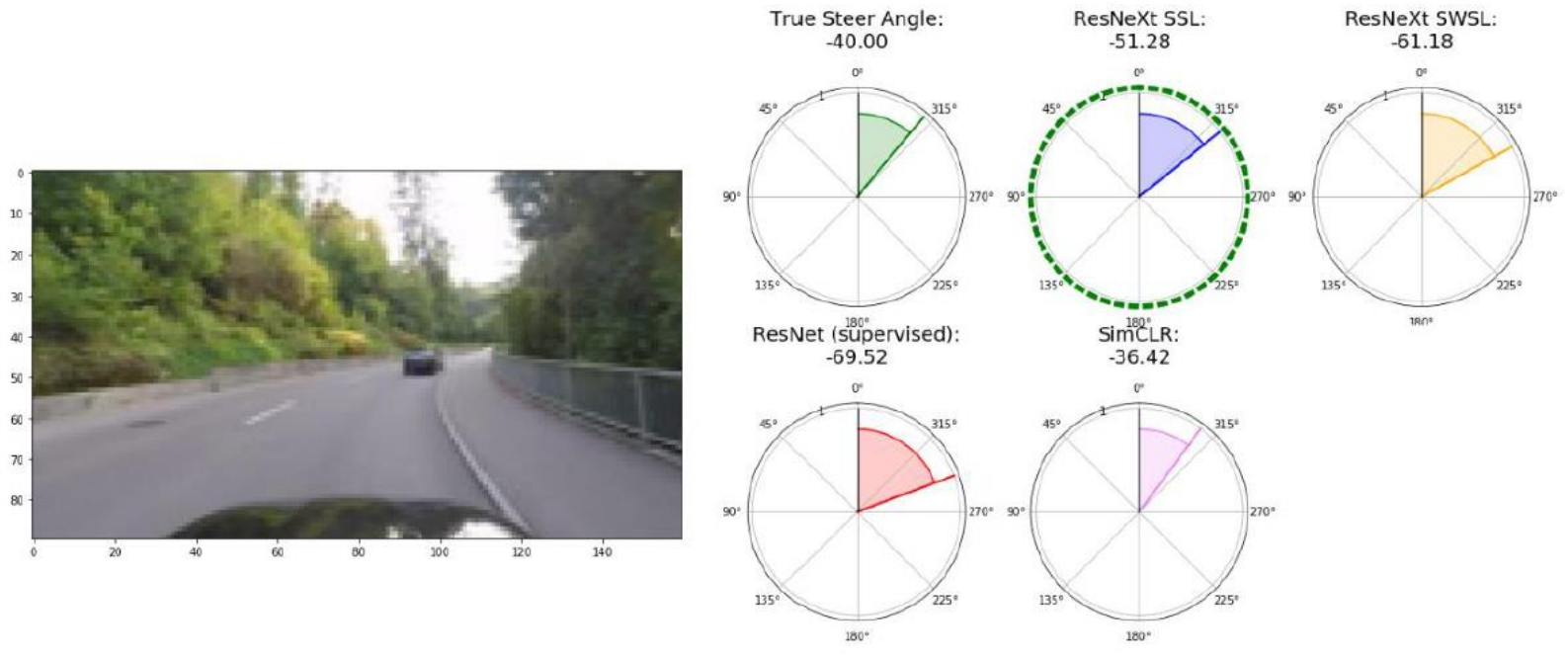}}

\subfigure{
\includegraphics[width=0.5\textwidth]{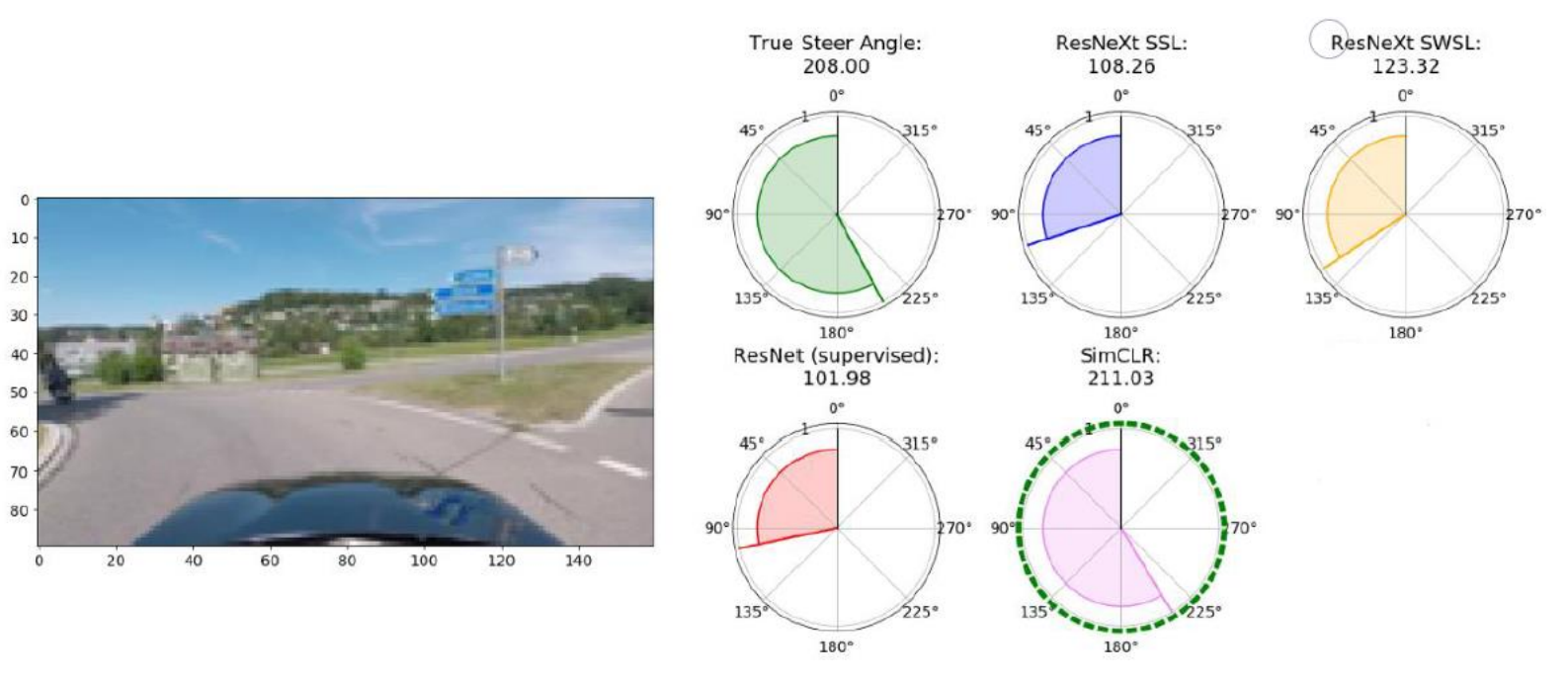}}
\caption{Steering wheel angle predictions from three examples in the Drive360 dataset. Using the low-level representation of the front-facing camera image (left), the model predicts the steering wheel angle 1 second in the future. To the right of each image, we show the ground truth in the top-right position, the prediction from the supervised ResNet-50 model in the bottom-left, and the predictions from the three semi-supervised models in the right positions. We highlight the most accurate model with a dashed green line. In all three examples, one of the semi-supervised models outperforms the supervised model.}
\label{fig:sampleDrive360}
\end{figure}

\bibliographystyle{splncs}
\bibliography{bibliography}

\end{document}